\documentclass[11pt]{article} % use larger type; default would be 10pt

\usepackage[utf8]{inputenc} % set input encoding (not needed with XeLaTeX)

\usepackage{authblk}

\usepackage{geometry} % to change the page dimensions
\usepackage{graphicx} % support the \includegraphics command and options

\usepackage{amssymb}
\usepackage{amsmath}
\usepackage{amsthm}
\usepackage{mathtools}
\usepackage{stmaryrd}
\usepackage{xfrac}
\usepackage{booktabs} % for much better looking tables
\usepackage{array} % for better arrays (eg matrices) in maths
\usepackage{paralist} % very flexible & customisable lists (eg. enumerate/itemize, etc.)
\usepackage{verbatim} % adds environment for commenting out blocks of text & for better verbatim
\usepackage{subfig} % make it possible to include more than one captioned figure/table in a single float
\usepackage{makecell}

\usepackage{fancyhdr} % This should be set AFTER setting up the page geometry
\usepackage{sectsty}
\allsectionsfont{\sffamily\mdseries\upshape} % (See the fntguide.pdf for font help)
\usepackage[nottoc,notlof,notlot]{tocbibind} % Put the bibliography in the ToC
\usepackage[titles,subfigure]{tocloft} % Alter the style of the Table of Contents

\title{Regularizing cross entropy loss via minimum entropy and K-L  divergence .}

\author{Abdulrahman O. Ibraheem \\ rahman.ibraheem1@outlook.com}

\date{}                     %% if you don't need date to appear

\begin{document}

\maketitle

\begin{abstract}
I introduce two novel loss functions for classification in deep learning. The two loss functions extend standard cross entropy loss by regularizing it with minimum entropy and Kullback-Leibler (K-L) divergence terms. The first of the two novel loss functions is termed mixed entropy loss (MIX-ENT for short), while the second one is termed minimum entropy regularized cross-entropy loss (MIN-ENT for short). The MIX-ENT function introduces a regularizer that can be shown to be equivalent to the sum of a minimum entropy term and a  K-L divergence term. However, it should be noted that the  K-L divergence term here is different from that in the standard cross-entropy loss function, in the sense that it swaps the roles of the target probability, $p(y|x)$, and the hypothesis probability, $\hat{p}(y|x)$. The MIN-ENT function simply adds a minimum entropy regularizer to the standard cross entropy loss function. In both MIX-ENT and MIN-ENT, the minimum entropy regularizer minimizes the entropy of the hypothesis probability distribution which is output by the neural network. Experiments on the EMNIST-Letters dataset shows that my implementation of MIX-ENT and MIN-ENT lets the VGG model climb from its previous 3rd position on the paperswithcode leaderboard to reach the 2nd position on the leaderboard, outperforming the Spinal-VGG model in so doing. Specifically, using standard cross-entropy, VGG achieves $95.86\%$ while Spinal-VGG achieves $95.88\%$ classification accuracies, whereas using VGG (without Spinal-VGG) our MIN-ENT achieved $95.933\%$, while our MIX-ENT achieved $95.927\%$ accuracies. The pre-trained models for both MIX-ENT and MIN-ENT are at https://github.com/rahmanoladi/minimum\_entropy\_project.
\end{abstract}

\section{Introduction}
Entropy is a well known measure of the randomness or disorderliness of a probability distribution. This means that sharp, and well-peaked distributions will have very low entropy, while flatter distributions will have higher entropy. Intuitively, this might have an implication for classification problems, where the task is to separate/classify inputs based on their underlying probability distributions. In particular, when the distribution of each class is sharp and well-peaked, the classification becomes less-prone to errors, whereas as the class distributions get flatter, the model gets prone to errors due to heavy tails and/or outliers. It is therefore reasonable to hypothesize that classification problems might benefit more when the class distributions have lower entropy. To this end, this work proposes two novel loss functions, namely MIX-ENT and MIN-ENT, which extend standard cross entropy loss by regularizing it with minimum entropy and Kullback-Leibler (K-L) divergence terms. Indeed, a similar idea has been explored with good results in \cite{min_ent_doc}, where they derived a \lq minimum system entropy\rq (MSE) criterion for document classification. Their MSE assigns an input document into the class that results in the least increase in overall system entropy. However, there are several differences between their work and the approach presented in this work. First, unlike theirs, we take a deep learning approach here, where for instance, we use state-of-the-art benchmark models like VGG-5 \cite{vgg_5}. Second, unlike theirs, our own approach builds on top of the time-honored cross entropy (CE) loss function, because we only add additional regularizers to the  CE loss function to arrive at our new loss functions. This approach has the benefit that the CE loss function can be recovered as a special case of our new loss functions, so that we can expect that the our loss functions should at least perform as well as the CE loss. Indeed, experiments on the EMNIST-Letters dataset \cite{emnist_letters} show that my implementation of  the new loss functions lets the VGG-5  model climb from its previous 3rd position on the paperswithcode leaderboard to reach the 2nd position on the leaderboard, outperforming the Spinal-VGG model \cite{kabir} in so doing. Specifically, using standard cross-entropy, VGG achieves $95.86\%$ while Spinal-VGG achieves $95.88\%$ classification accuracies, whereas using VGG (without Spinal-VGG) our MIN-ENT achieved $95.933\%$, while our MIX-ENT achieved $95.927\%$ accuracies. The pre-trained models for both MIX-ENT and MIN-ENT are at https://github.com/rahmanoladi/minimum\_entropy\_project.

\section{Description of the first proposed loss function (MIX-ENT)}
The first proposed criterion, namely MIX-ENT, regularizes CE loss with a term that can be shown to be equivalent to the sum of a minimum entropy regularizer and a Kullback-Leibler regularizer. To see how this is the case, suppose that $x$ is an input to a neural net, and that $y$ is the label associated with $x$. Further, let $p(y|x)$ denote the true/target probability distribution which the neural net is trying to learn, and let $\hat{p}(y|x) $ be the neural net's output; we will also refer to $\hat{p}(y|x) $ as the hypothetical probability distribution. The standard cross entropy, herein denoted as $\mathcal{L}_{CE1}$, can be written as:

\begin{equation}
\label{eqn_1}
\mathcal{L}_{CE1} =  -  \sum_{y} p(y|x)log( \hat{p}(y|x)) 
\end{equation}

Now, we will derive a well known result which simply says that cross entropy is sum of KL-divergence and entropy. We  begin with:  

\begin{equation}
\label{eqn_2}
 -  \sum_{y} p(y|x)log( \hat{p}(y|x)) =   - \sum_{y} p(y|x)log( \hat{p}(y|x)) + \sum_{y} p(y|x)log( p(y|x)) - \sum_{y} p(y|x)log( p(y|x))
\end{equation}  

\begin{equation}
\label{eqn_3}
 -  \sum_{y} p(y|x)log( \hat{p}(y|x)) =    \sum_{y} p(y|x)log( \frac{p(y|x)) } { \hat{p}(y|x)} ) - \sum_{y} p(y|x)log( p(y|x))
\end{equation}

Now observe that the first quantity, namely $- \sum_{y} p(y|x)log( \hat{p}(y|x)) + \sum_{y} p(y|x)log( p(y|x))$,  on the R.H.S. of Equation \ref{eqn_3} above is the the K-L divergence, using $p(y|x)$ as the weighting probability distribution. Also, observe that the second quantity, namely $- \sum_{y} p(y|x)log( p(y|x))$, on the R.H.S. of the equation is just the entropy of the target distribution, $p(y|x)$. This entropy is denoted as $H(p)$, and this leads to the following intuition: if it were the case that entropy $H(p)$ could be varied, then we would have been able to minimize entropy by minimizing cross entropy. But, in reality $H(p)$ is fixed, because it is the entropy of the target probability distribution, so we can not hope to minimize it. But, this analysis reveals that if we had used $\hat{p}$ in place of $p$ in our definition of cross entropy above, then we would have arrived at a loss function which allows us to minimize $H(\hat{p})$ instead of $H(p)$. Now, this observation just perfectly fits into the objective we started out with: to teach the neural net how to minimize the entropy of its hypothetical distribution. To show this clearly, we will define a \lq swapped cross entropy\rq criterion, $\mathcal{L}_{CE2}$ which simply swaps the roles of $p(y|x)$ and  $\hat{p}(y|x)$ in  the expression of $\mathcal{L}_{CE1}$ above. We have:

\begin{equation}
\label{eqn_4}
\mathcal{L}_{CE2} =  -  \sum_{y} \hat{p}(y|x)log( p(y|x)) 
\end{equation}

\begin{equation}
\label{eqn_5}
 -  \sum_{y} \hat{p}(y|x)log( p(y|x)) =   - \sum_{y} \hat{p}(y|x)log( p(y|x)) + \sum_{y}\hat{p}(y|x)log(\hat{p}(y|x))- \sum_{y}\hat{p}(y|x)log(\hat{p}(y|x))
\end{equation}

Next, we have:
\begin{equation}
\label{eqn_6}
 -  \sum_{y} \hat{p}(y|x)log( p(y|x)) =    \sum_{y} \hat{p}(y|x)log( \frac{\hat{p}(y|x)) } {p(y|x)} ) - \sum_{y} \hat{p}(y|x)log( \hat{p}(y|x))
\end{equation}

The first term on the R.H.S. of Equation \ref{eqn_6} above is a KL divergence which is weighted by the the hypothetical distribution, $\hat{p}(y|x)$. We will denote it as $KL_{\hat{p}}$. Note that this KL divergence term is quite different from the KL divergence in Equation \ref{eqn_3} which was weighted by the target distribution. Further, the second term on the R.H.S. of Equation \ref{eqn_6} above is entropy, $H(\hat{p})$, of the hypothetical distribution. Consequently, by minimizing  $\mathcal{L}_{CE2}$, we would be simultaneously minimizing $KL_{\hat{p}}$ and $H(\hat{p})$. Tellingly, minimizing $KL_{\hat{p}}$ and $H(\hat{p})$ should be good, because minimizing $KL_{\hat{p}}$ can be viewed as making the hypothetical distribution align as much as possible with the target distribution, while minimizing  $H(\hat{p})$ makes the hypothetical distribution as sharp and well-peaked as possible. Based, on the foregoing, the proposed MIX-ENT minimizes a weighted linear combination of  $\mathcal{L}_{CE1}$  and $\mathcal{L}_{CE2}$. We spell it out thus:

\begin{equation}
\label{eqn_7}
\mathcal{L}_{MIX} =  \beta_1 \mathcal{L}_{CE1}  + \beta_2 \mathcal{L}_{CE2}  
\end{equation} 
 
Above $\beta_1 $ and $\beta_2$ are learnable scalars which are learnt along with the model parameters, but with possibly different learning rates. Finally, we define the second proposed loss function, MIN-ENT as follows:

\begin{equation}
\label{eqn_7}
\mathcal{L}_{MIN} =  \beta_1 \mathcal{L}_{CE1}  + \beta_2 H(\hat{p})
\end{equation} 

That is, MIN-ENT simply regularizes standard cross entropy with the minimum entropy term. Finally, in both MIX-ENT and MIN-ENT the bases of the logarithms are also learnt along with the model paramaters.

\section{Experiments and Results}

I conducted experiments on the EMNIST-Letters dataset, using the VGG-5 as base model, to compare the proposed loss functions with standard cross entropy loss.  Using Ray Tune's HyperOpt searcher and ASHA scheduler\cite{ray_tune}, I first performed an hyper-parameter sweep to find optimal hyper-parameters, such as batch-size, learning rate, weight decay and dropout  for the baseline case of standard cross-entropy. This hyper-parameter sweep involved 1000 trials with a maximum of 50 epochs per trial. I then took the best model returned by the search, and trained it for a further 150 epochs. A similar proceudre was carried out for the MIX-ENT and MIN-ENT loss functions. Results are in Table 1. In addition to the results from my own experiments, I have also included results from the paperswithcode leaderboard. From the table, we see that the MIX-ENT and MIN-ENT criteria do not only outperform my own results for standard cross entropy, they also outperform the Leaderboard's results for standard cross entropy. Consequently, this puts VGG trained with MIN-ENT and MIX-ENT on the second and third positions on the leaderboard respectively. ( since MIN-ENT slightly outperforms MIX-ENT). More importantly, as can be seen from the leaderboard results shown in Table 1, the results also puts the VGG model trained with MIX-ENT and MIN-ENT ahead of the more sophisticated Spinal-VGG model trained with standard cross-entropy.

 \begin{table}[htb]
\caption{Comparison of MIX-ENT and MIN-ENT against cross entropy on EMNIST-Letters dataset}
\centering

\resizebox{\columnwidth}{!}{%
\begin{tabular}{ccccc}
\toprule
 \makecell[t]{Back-bone\\ model}& \makecell[t]{loss\\function}& \makecell[t]{Final epoch\\ accuracy} &  \makecell[t]{Best epoch \\accuracy} &\makecell[t]{ Best epoch\\ Leaderboard accuracy}\\
\midrule
                           VGG                  &    Cross entropy                                         &    95.322                               & 95.707&   95.860 \\
                           Spinal VGG                  &    Cross entropy                                         &    -                               & -&   95.880 \\
                           VGG                  &    MIX-ENT                                         &    \textbf{95.870}                               & \underline{95.928}&  \underline{95.928}  \\
                           VGG                  &   MIN-ENT                                &    \underline{95.865}                               & \textbf{95.933}&  \textbf{95.933}\\
\end{tabular}%
}

\end{table}

\section{Conclusion}
I have introduced two novel loss functions, called MIX-ENT and MIN-ENT, which extend standard cross entropy by regularizing it with a minimum entropy term and a KL-divergence term. The intuition is that minimizing entropy should sharpen the learnt probability distribution, and make it more peaked. I compared the proposed loss functions with standard cross-entropy on the EMNIST-Letters dataset using the VGG-5 as a base model. Results show that the proposed loss functions outperform standard cross entropy on the EMNIST-Letters dataset, and that the results infact put VGG-5 trained with MIN-ENT and MIX-ENT on the second and third positions on the leaderboard respectively, ahead of the more sophisticated Spinal-VGG trained with standard cross entropy loss.

\end{document}